# NeuralSentinel: Safeguarding Neural Network Reliability and Trustworthiness[*]

Xabier Echeberria-Barrio, Mikel Gorricho, Selene Valencia, and Francesco Zola

Vicomtech Foundation, Basque Research and Technology Alliance (BRTA);
Paseo Mikeletegi, 57, Donostia 20009, Spain
e-mail: {xetxeberria, mgorricho, svalencia, fzola}@vicomtech.org.

**Abstract.** The usage of Artificial Intelligence (AI) systems has increased exponentially, thanks to their ability to reduce the amount of data to be analyzed, the user efforts and preserving a high rate of accuracy. However, introducing this new element in the loop has converted them into attacked points that can compromise the reliability of the systems. This new scenario has raised crucial challenges regarding the reliability and trustworthiness of the AI models, as well as about the uncertainties in their response decisions, becoming even more crucial when applied in critical domains such as healthcare, chemical, electrical plants, etc. To contain these issues, in this paper, we present NeuralSentinel (NS), a tool able to validate the reliability and trustworthiness of AI models. This tool combines attack and defence strategies and explainability concepts to stress an AI model and help non-expert staff increase their confidence in this new system by understanding the model decisions. NS provide a simple and easy-to-use interface for helping humans in the loop dealing with all the needed information. This tool was deployed and used in a Hackathon event to evaluate the reliability of a skin cancer image detector. During the event, experts and non-experts attacked and defended the detector, learning which factors were the most important for model misclassification and which techniques were the most efficient. The event was also used to detect NS's limitations and gather feedback for further improvements.

## 1 Introduction

The interconnection between Information Technologies and Operational Technologies is underway, with many impacts on the related cybersecurity of various application domains. In fact, in this new scenario, attacks or malfunctions in the cyber world can have critical impacts on the physical world, especially in critical domain such as healthcare, chemical, biological, electrical plants, or nuclear ones. In this interconnected cyber-physical world, the advent of Artificial Intelligence (AI) opens the door to implement and improve numerous defence capabilities. Indeed, from a traditional point of view, these AI models are mainly used for supporting users, practitioners and experts in anomaly detection and classification tasks, i.e., in detecting attacks, malfunctions, or anomaly behaviours that can erode the system functionality. These applications help to reduce the human efforts across their entire workflow.

[*] This work has been partially supported by the European Union's Horizon Program under the project KINAITICS (Grant Agreement No. 101070176). Views and opinions expressed are, however, those of the author(s) only and do not necessarily reflect those of the European Union. Neither the European Union nor the granting authority can be held responsible for them.

Several studies [1] state that even without a deep understanding of AI, humans in the loop can build trust with an AI system through experience, expert endorsement and validation. However, it is important to highlight that the usage of these AI models led them, in turn, to become system attack points that can compromise the reliability of the ecosystems. In fact, hackers and intruders can directly attack these AI models in order to change their predictions, and thus skew the human decision. In literature, many AI attacks and defence techniques are studied and compared, with the aim of raising such problems. This situation raises crucial challenges regarding trustworthiness and dealing with uncertainties in response decisions, becoming even more crucial when critical infrastructures are involved. In fact, a human decision taken on a skewed reality in these environments can greatly impact the entire ecosystem and endanger the human lives involved.

To contain these issues, once a new AI system is implemented, before its application in the real world, it needs to be stressed and attacked - but also opportunely defended - to check its reliability and trustworthiness. This operation will also help humans in the loop to be conscious of possible AI malfunctions, how they are generated, what they need to look at carefully, and how they can affect the model results.

For this reason, in this paper, we present NeuralSentinel (NS), a tool able to validate the reliability and trustworthiness of Artificial Neural Networks (ANN) models. NS give the opportunity to the human in the loop to attack and defend a loaded ANN model with an easy and self-explanatory user interface. The tool allows the application of different state-of-the-art attack and defence strategies, that can also be combined in order to detect the most suitable one for the considered use case. Furthermore, the interface gives the opportunity to show the raw input data, the attacked data and a difference computed between the two sets. This difference highlights the points that can generate the misclassification in the model. These points usually are not visible to humans directly in the attacked data. Finally, NS includes an explainability (XAI) module that helps users understand the model prediction. This XAI module aims to approach non-expert staff to understand the model decisions based on the model inputs.

NS has been validated during a Hackathon event organized in the KINAITICS European project[1]. In particular, it was applied to attack and defend an AI model related to a Healthcare Use Case, i.e., a model trained for detecting skin cancer from skin images. During the event, on the one hand, hackathon users learned more about the reliability and trustworthiness of an AI decision-making system. On the Other hand, we obtain feedback for improving the functionality and the usability of the tools itself. To the best of our knowledge, NeuralSentinel is among the first tools that allow users to combine several attack/defence strategies and also provide an XAI module to help them acquire confidence in the model.

---
[1] https://kinaitics.eu/

The rest of the paper is organized as follows. Section 2 introduces common attack and defence strategies on AI models, explainability concepts and reports related work. After that, Section 3 presents the NeuralSentinel tool, its architecture, and its interface. Then, the validation results obtained in a particular use case are reported in Section 4. Finally, conclusions and guidelines for future work are drawn in Section 5.

## 2 Preliminaries

### 2.1 Attack on Artificial Intelligence

Artificial Neural Networks are sensitive to minor modifications according to the biases they can contain. Adversarial attacks, specifically adversarial examples, exploit those biases by introducing imperceptible noise into input data, leading to erroneous outputs[2]. In this context, *noise* refers to imperceptible perturbations or alterations introduced into input data. These perturbations are strategically applied to confuse or deceive neural networks, causing them to produce incorrect or undesired outcomes. The *noise* is essentially a manipulation of input data that is typically very subtle and difficult to detect by the human eye. Nevertheless, it can significantly impact the neural network's performance by causing it to make erroneous predictions or classifications [3]. Adversarial attacks often exploit neural networks' sensitivity to such perturbations, making them particularly challenging and problematic, especially in applications demanding high precision and reliability, such as medical diagnosis or autonomous driving safety[4].

Adversarial attacks can compromise the effectiveness of early disease detection or the safety of automated systems. Thus, the development of robust defences against these threats is critical to secure the continued success of neural networks in these vital applications [5]. The inception of adversarial examples in deep neural networks can be traced back to the early work of Szegedy et al. [6], who utilized the Limited-Broyden Fletcher Goldfarb Shanno (L-BFGS) algorithm to generate the first documented adversarial example. Subsequently, the field has witnessed the development of more efficient and less detectable algorithms for adversarial example generation. Therefore, there exist several different algorithms to obtain an adversarial example of the targeted model:

 1. **Fast Gradient Sign Method (FGSM):** In 2014, Goodfellow et al. [7] presented the Fast Gradient Sign Method (FGSM) algorithm. FGSM involves calculating the gradient of the neural network's loss function with respect to the input data and then making small adjustments to the input in the direction that maximizes the loss. This perturbation is typically very subtle but can lead to the misclassification of data by the neural network, highlighting its vulnerability to adversarial attacks.

2. **Basic Iterative Method (BIM):** The Basic Iterative Method (BIM) algorithm [8] is an extension of the FGSM algorithm. BIM iteratively applies small perturbations to the input data and gradually increases the magnitude of these perturbations over multiple iterations. This method aims to find a more effective adversarial example by iteratively adjusting the input data in the direction that maximizes the loss function.

3. **Projected Gradient Descent Method (PGD):** The Projected Gradient Descent Method (PGD) algorithm [9] builds upon the BIM algorithm. PGD applies iterative updates to input data while ensuring that the perturbed data remains within an epsilon-sized perturbation ball around the original input. This method performs gradient descent on the loss function, adjusting the input data step by step, with the constraint of staying within a bounded region.

4. **Jacobian-based Saliency Map Attack (JSMA):** In 2016, Papernot et al. [10] introduced the Jacobian-based Saliency Map Attack (JSMA) algorithm. JSMA works by saturating the pixels of an image to the maximum or minimum level. In this way, JSMA can cause the model to misinterpret the input, in this case, the image, giving an erroneous output or prediction. Moreover, Wiyatno et al. developed two new variants of this attack in [11].

5. **DeepFool algorithm:** In 2016, Moosavi-Dezfooli et al. [12] presented the DeepFool algorithm which is based on a gradient-free approach. It operates by iteratively estimating the decision boundary of a neural network and computing the minimum perturbation required to move an input data point across that boundary. The core idea behind DeepFool is to find the closest decision boundary of a neural network by measuring the linear distance between the input and the boundary. It then perturbs the input in the direction that minimizes this distance. The attack continues iteratively until the network classifies the perturbed input incorrectly.

## 2.2 Defend Strategies

Mitigating adversarial attacks in neural networks revolves around developing defensive measures to reinforce a network's robustness and its ability to maintain reliable performance even in the presence of deliberate, subtle perturbations in input data. These measures aim to safeguard the integrity of neural network systems against potential threats that could exploit vulnerabilities within the models.

Researchers and practitioners are deeply engaged in the quest for ways to bolster the network's innate resistance against adversarial manipulations. The overarching goal is to ensure that neural networks can continue to function correctly and provide reliable outputs in real-world applications despite the ever-evolving landscape of adversarial techniques. The following section provides an overview of the proposed countermeasures:

1. **Adversarial training:** The technique of adversarial training is first introduced in [6]. Adversarial training involves training the neural network on a combination of clean (unaltered) data and adversarial examples generated during the training process. The key idea is to expose the model to these corrupted samples to help it learn to better withstand and classify them correctly. During adversarial training, the model is continually tested with perturbed input data, and it learns to adapt its parameters to be more resilient to adversarial perturbations. This process can result in a more robust model that can better handle real-world scenarios where adversarial attacks might be attempted.

2. **Dimensionality Reduction:** This defence technique was presented in [13]. The dimensionality reduction method is a powerful technique that effectively leverages autoencoders to reduce noise in input data. This work presented two approaches to dimensionality reduction within adversarial defences, demonstrating resilience against potential attacks. The first approach involves directly applying an autoencoder to the input data. In this configuration, the autoencoder is tasked with learning a lower-dimensional representation of the data that captures its essential features while minimizing the noise contained in the input data. Concretely, the perturbation introduced by the adversary can be considered noise, and hence, the dimensionality reduction works as a filter of this perturbation, reducing the introduced small modifications. Therefore, this method makes the target model more difficult to exploit by adding perturbations in the input data. The second approach integrates the autoencoder as an intermediary component within the neural network architecture. Here, the autoencoder is strategically positioned between the input and the primary layers of the network. It follows the same idea as the previous one, but considering that the perturbation introduced in the input data will modify the embedding returned by the feature extractor part of the target model. The incorporated perturbation will be reflected as noise in the embedding, and hence, the dimensionality reduction will work in the same way but at the embedding level. Highlight that this method preserves the original neural network's integrity. The weights and architecture of the network remain unaltered throughout the dimensionality reduction process, ensuring that the core functionality and learned representations of the model are maintained.

3. **Prediction Similarity:** The prediction similarity defence was presented by Echeberria-Barrio et al. [13]. In this case, an external layer is also applied to the model, so it does not modify the model to be protected. This method aims to place a layer in the data input to detect possible attempts of adversarial attacks.

This layer functions as a repository for recording the historical data of inputs, predictions, and specially crafted features. These features draw inspiration from the concept that adversarial attacks often rely on multiple predictions of similar images to create adversarial examples. The information stored within this layer can be leveraged to create a risk assessment feature, enabling the evaluation of whether the input is indicative of an adversarial attempt or not.

4. **Defensive distillation:** In 2016, Papernot et al. presented a defensive dilatation technique in [14]. Defensive distillation is one of the early methods aimed at improving the security of neural networks, particularly in the context of adversarial attacks. The key idea behind defensive distillation is to train the 'student' model (the original neural network) using the softened probability distributions produced by the teacher model instead of the true labels from the training data. The teacher model acts as a source of knowledge and transfers information about the data distribution to the student model. This results in a network that generalizes better and is less sensitive to adversarial perturbations, as the soft targets provide a smoother and more robust loss landscape during training.

5. **Gradient masking:** Lee et al. introduced a defence based on gradient masking in [15]. It occurs when the gradients of the network's loss function concerning the input data become too small or vanish as the network's layers are traversed during backpropagation. Regarding adversarial attacks, gradient masking is problematic because it makes it difficult to obtain meaningful gradient information to craft effective input data perturbations. When gradients are masked, an attacker may not be able to identify the most sensitive features in the network or the direction in which they should perturb the data to deceive the model.

## 2.3 Explainability and Interpretability

Explainability and interpretability in AI are methods to make black-box AI models more understandable. Concretely, the interpretability methods focus on why and how the AI models work, while the explainability techniques try to explain their decisions. Both help ensure the secure development of AI systems, allowing users to have confidence in the decisions made by these models and enabling them to clarify and identify vulnerabilities or potential threats.

Some machine learning models, like linear models, offer inherent interpretability. In a linear model, predictions are derived from a straightforward equation ($Y = aX + b$), resembling a simple straight line on a plot. The parameters $a$ (feature weight) and $b$ (intercept) are easily understood, making linear models user-friendly. However, more sophisticated machine learning models with millions of neurons, such as neural networks, become necessary to obtain more complex predictions or classifications from a dataset. In this context, various explainability and interpretability methods, including visualization methods, have been explored.

SHapley Additive exPlanations (SHAP)[16] is a technique based on game theory and attributes feature contributions to a global and local prediction. In the context of SHAP, each feature is considered a player in a cooperative game, and the impact of each feature on a prediction is calculated considering all possible combinations of features. Local Interpretable Model-Agnostic Explanations LIME [17] is another interpretability technique that provides local explanations for model predictions by selecting specific data, introducing small random feature modifications and finally evaluating the ML model on this feature samples. This interpretable model captures how features relate to predictions in the neighborhood of the selected instance. Another example of interpretability method can be Learning Important FeaTures (DeepLIFT) [18], an attribution technique for understanding how input features contribute to predictions made by deep neural networks. Its main focus is calculating the relative contributions of features across neural network layers.

In addition to the techniques mentioned above, visualizations play a critical role in the interpretability of Machine Learning models. Data and model predictions are often inherently complex and challenging to understand without visual representations. Visualizations, such as bar charts, heat maps, and scatter plots, provide an intuitive representation of data and relationships between features. These visualizations can help identify data patterns, trends, and relationships, which is crucial for model interpretation.

An example of visualization for interpretability can be Deconvolution[19]. Deconvolution, also known as transposed convolution or fractionally strided convolution, is a mathematical operation often used in deep learning and computer vision. Deconvolution can help generate feature maps that maximize the activation of a particular neuron or layer within the network. This method allows users to visualize what the network has learned to detect. Class Activation Map (CAM)[20], is another interpretability technique used in the field of computer vision, especially in convolutional neural networks (CNN). It shows which parts of an image are most relevant for the model to perform a specific classification. Friedman et al. [21] introduced Partial Dependence Plots (PDPs) as a visualization tool designed to aid in interpreting black-box predictive models. By isolating a specific feature and systematically varying its values while keeping other variables constant, the PDPs tool helps to understand how that feature impacts the model's output. Similar to PDP, Individual Conditional Expectation (ICE)[22], is an interpretative technique in machine learning. ICE provides insights into the effect of a single feature on the predictions of a model by visualizing how that feature's value influences the model's output for individual data points.

The explainability method for artificial neural networks used in this paper was proposed by Echeberria-Barrion et al. in [23].. Their approach analyses the behavior of the targeted model, monitoring the activation of the main nodes. The method consists of first detecting the most frequent nodes in each class of the model, i.e., those that are activated most frequently in the images of each class. This method contains a preprocess that starts calculating the frequencies of the model's neurons for each class. It keeps comparing the frequency associated with a neuron in a class with the frequencies of the same neuron in the rest of the classes. This process is repeated for all the neurons in the model. In this way, each class is compared with all the others. The comparison is made by subtracting the frequencies of the neuron activations in absolute value. Once the preprocess is computed, this method is ready to be implemented. Suppose the targeted attack is modifying an image with the original class $c$ to convert it to class $c'$. In this case, the method takes the comparison between class $c$ and $c'$ and takes $k$ neurons with the highest frequency difference. Those top $k$ neurons are monitored to see the effect of the attack on the model.

## 3 NeuralSentinel

NeuralSentinel (NS) is a powerful tool designed to assess the reliability and trustworthiness of Artificial Neural Network (ANN) models. NS empowers human users to both test and safeguard loaded ANN models. Furthermore, it enables the implementation of various cutting-edge attack and defence strategies, which can also be customized and combined to identify the optimal approach for a specific use case.

The presented tool is divided into three main modules: NeuralSentinel, API, and Visualization. Concretely, Figure 1 shows how these modules are connected and operate. Furthermore, Figure 1 shows two pipelines: the *initial* one, processes in the dotted blue line, and the *execution* one, processed in the dotted red line.

The *initial pipeline* includes the operations that NS needs at the beginning to instantiate (load) a model and data to carry on all the implemented functionalities. Therefore, once the initial model and data are given, the user can exploit all the NS functionalities in the *execution pipeline*.

All the NS modules are further detailed in the next sections.

### 3.1 NeuralSentinel module

The NeuralSentinel module represents the back-end of the NS tool. It contains three main submodules: Attack, defence, and Explainability (XAI). However, to perform its operations, this module requires that the model and their related data are loaded, i.e., the initial pipeline is completed. In particular, the Attack submodule allows users to deploy three of the main attacks discussed in Section 2.1, which are FGSM, BIM, and PGD. At the same time, the defence submodule contains three main defence mechanisms such as adversarial training, dimensionality reduction, and prediction similarity (Section 2.2). Finally, the XAI submodule includes the main node activation monitoring explainability method. To do that, it requires that the module and its data are correctly loaded for then initialize the main node activation monitoring explainability method described in Section 2.3. Therefore, the XAI calculates the frequency difference of the nodes in the model by classes. In other words, it takes data, computes the activation frequency of each node in all the classes (grouping the data in the dataset by classes), takes the class *c*, compares this frequency with the frequencies in all the rest of the classes, and saves those numbers as the importance of the nodes in the class *c*. After this initialization, the NeuralSentinel module is ready to receive any petition implemented.

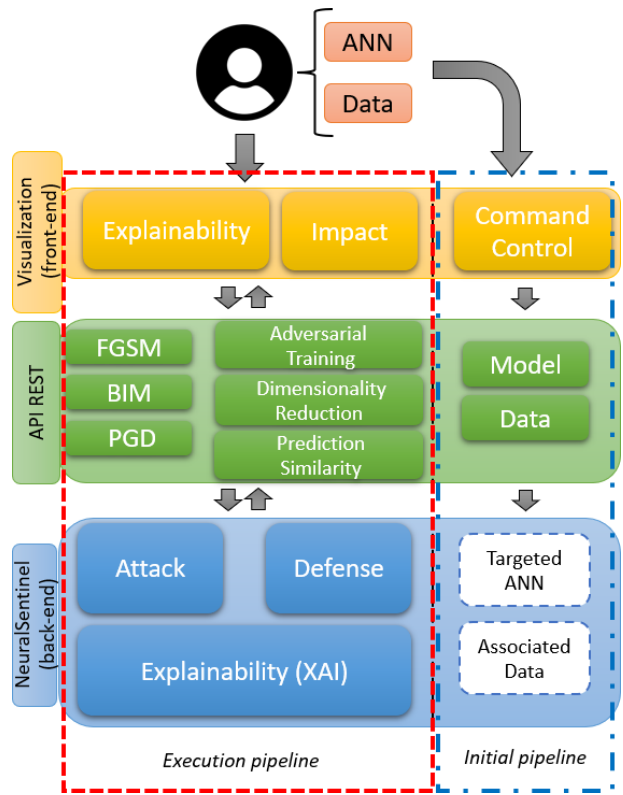

Fig. 1: NeuralSentinel architecture

### 3.2 API module

The API REST module is in charge of connecting the Visualization module with the NeuralSentinel module, i.e., to connect the front-end and the back-end of the application. In particular, it helps the user to perform complex operations related to attack and defence activities in an easy way.

The API module enables three main functionalities: loading models and data, evaluating an attack and evaluating a defence mechanism. The first call is self-explanatory and it allows users to select a local model and data to be imported into the server. The second functionality allows the users to modify a chosen attack algorithm as they need, i.e., they can define several parameters for changing the impact and the effectiveness of the attack. In that sense, the FGSM attack requires one parameter ($\epsilon$), while the BIM and the PGD attacks require two parameters ($\epsilon$ and steps). Once played, they take the selected model and some of the samples in the data for their actions. Finally, the third functionality allows users to evaluate the defence mechanism. To do that, they must select and define a concrete attack's name and their respective parameters, as mentioned above, and then they also need to select a specific defence mechanism among the ones available (adversarial training, dimensionality reduction, and prediction similarity).

### 3.3 Visualization module

The visualization interface allows the user to execute the operations and visualize the results. In particular, it is composed of three main views: *Command Control, Impact* and *Explainability*. The *Command Control view* (Figure 2) allows the user to load - or select a preloaded - model to perform the analysis through a drop menu and a simple button. The platform also allows users to check more information about the model itself (if this information is opportunely preloaded). Then, users can select through radio buttons which kind of adversarial attack they want to deploy over the model and which defence mechanisms should be used to protect it. According to the chosen attack, users must define several parameters shown in numeric text boxes, since they are the key for the attack (as described in Section 2.1).

Fig. 2: NeuralSentinel command control view

The *Impact view* reports the results gathered from the model, once the attack and defence mechanisms are deployed (Figure 3). In particular, this section displays the original and the attacked data, as well as the difference between them, in order to highlight the changes caused by the adversarial attack. In the example provided in Figure 3, images are used to show the usefulness of this visualization. Furthermore, the impact view includes a table that reports the model's accuracy with the original and the attacked sample, the simulation minimum, maximum and, average, and finally, the grade of the adversarial attack.

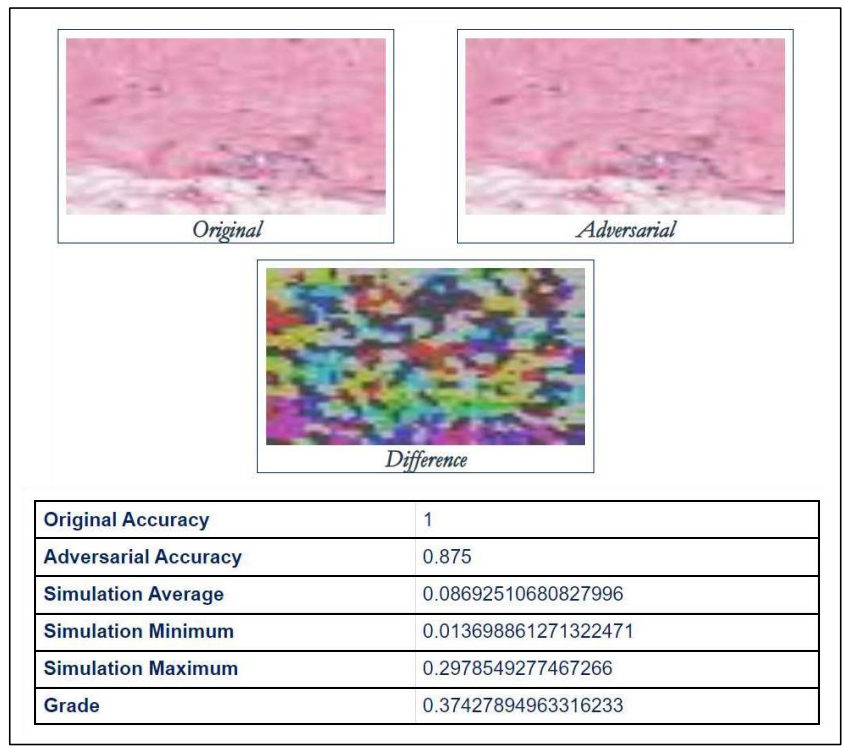

Fig. 3: NeuralSentinel impact view

- *Original accuracy:* It is the model's precision in the data taken for the attack. It helps to check the initial model's performance.
- *Adversarial accuracy:* It shows the model's precision in the modified data. The proportion of successful attacks can be checked by combining this information and the original accuracy metric. Therefore, this metric helps to understand the success of the implemented attack.
- *Similarity average:* This metric is the mean of comparing each modified data with its original. This comparison is computed using the SSIM similarity metric presented in [24]. It helps to understand how equal are the modified data, assuming that the more similar an adversarial example is, the better it is.

- *Similarity maximum*: This metric is the maximum similarity difference computed between the modified data and their original. It gives auxiliary information to the similarity average metric since it shows if some huge modifications are perturbating the average.
- *Similarity minimum:* This metric is the minimum similarity difference computed between the modified data and their original. It gives auxiliary information to the similarity average metric since it shows if some little modifications are perturbating the average.
- *Grade:* This metric gives a notion about the balance between perturbation and misclassification in adversarial examples generation. A corrupted image with huge noise can not be considered a 'good' adversarial example since it will be distinguishable for humans. At the same time, a tamper that does not get misclassification is useless. Moreover, both perturbation and misclassification are related. A more significant noise gives more options to obtain the misclassification, but the modification can generate a corrupted image distinguishable for humans. The metric ranges between 0 and 1, where 1 is the excellent evasion method, and 0 is the awful adversarial technique according to the mentioned balance.

Lately, the *Explainability view* allows users to understand better what happened in the model, and why its behaviours changed (Figure 4 and Figure 5). In particular, the view shows how the neurons act, i.e. their behaviours, in the model during the sample classification. Indeed, along with the attack process, NS monitors the activation value of the model's neurons. Once the attack is finalized, the XAI module uses randomly selected data (the same shown in the Impact view), to calculate and report the most interesting models' neurons (table in Figure 4), following the preprocess explained in Section 3.1. Remember that neurons are sorted by class comparisons. Therefore, suppose that the original class of the selected input data is $c$, and the implemented attack converts it into class $c'$. Then, the chosen neurons will be the top $k$ in the comparison between class $c$ and class $c'$. The results of this monitoring are shown since they help to detect a change in neurons' behavior and understand the model decision. Furthermore, the impact value of each neuron is displayed over the chosen attack steps, as shown in the graphs in Figure 5. This approach also helps reduce the model's attack surface and detect which part of the model itself is the most critical and needs more protection.

## 4 Validation

NS tool was used and validated during a Hackathon event organized in the KI-NAITICS European project. On the one hand, during this event, project end users, external practitioners, students and experts had the chance to test the functionality of the tool in a realistic use case. On the other hand, we had the opportunity to gather feedback from them in order to improve the tool and its usability.

| Neuron | Class 0 frequency (F0) | Class 1 frequency (F1) | Frequency difference (F0-1) |
|---|---|---|---|
| 28 | 0.7279 | 0.1958 | 0.5320 |
| 226 | 0.8659 | 0.3365 | 0.5293 |
| 44 | 0.5916 | 0.1438 | 0.4477 |
| 486 | 0.8349 | 0.4002 | 0.4346 |
| 124 | 0.5274 | 0.1325 | 0.3948 |
| 435 | 0.4329 | 0.0568 | 0.3760 |
| 254 | 0.2820 | 0.6419 | 0.3599 |
| 40 | 0.0700 | 0.4210 | 0.3510 |
| 265 | 0.1691 | 0.4804 | 0.3112 |
| 76 | 0.1486 | 0.4517 | 0.3030 |

Fig. 4: Explainability view: Neuron classification relevance by class frequency.

### 4.1 Healthcare Use Case

In the KINAITICS event, NS was used to evaluate the reliability and trustworthiness of a model trained for detecting cancer in skin breast images. To make more realistic the use case, and to emulate a healthcare application, a visualization server has been deployed. Specifically, a Digital Imaging and Communication in Medicine (DICOM) server is implemented in order to simulate the central point in which multiple frames or images are stored, containing a wealth of metadata, patient name, serial number, etc., as shown in Figure 6. DICOM is a standardized protocol for transmitting medical images and associated data between medical hardware. For its implementation, the popular open-source tool Orthanc[2] has been used. The DICOM was populated with a breast skin dataset[3].

As described in many studies [23][13], the common architecture for models involved in the diagnosis (cancer or no cancer) from skin breast images, involve the combination of VGG16 [25] (in some cases replaced with Convolutional Neural Networks CNNs) and a Dense Neural Network (DNN) layer. Following this trend, we trained a model with the same structure. Therefore, the DICOM data and the model itself represented the information that the NS tool needed in its *initial pipeline*. For this reason, during the Hackathon event, users are directly involved in the *execution pipeline*.

---
[2] https://www.orthanc-server.com/
[3] https://www.kaggle.com/datasets/paultimothymooney/breast-histopathology-images

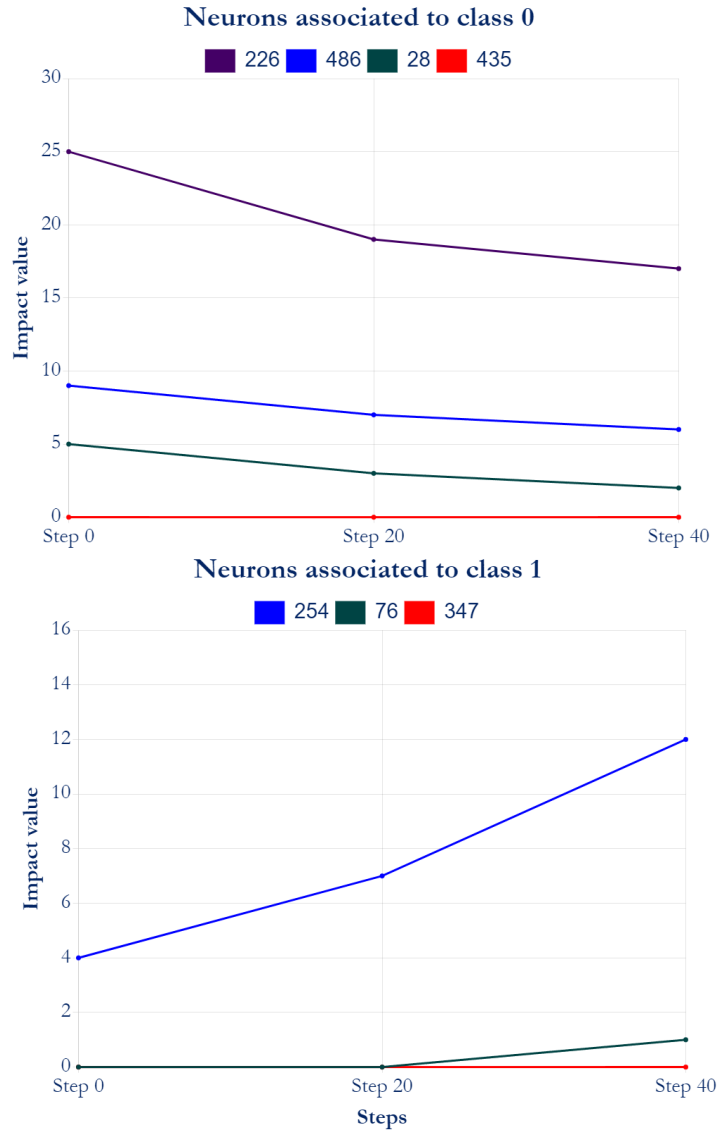

Fig. 5: Explainability view: Neuron impact value on the classification over increasing steps

### 4.2 Challenge Definition

The Hackathon event was organized as a Capture-the-Flag (CtF) exercise. CtF is a type of information security contest where participants are challenged to solve a range of tasks in order to obtain a designated item called a flag. In particular, there are many formats and strategies that can be implemented [26]. However, in the KINAITICS project, the *Jeopardy-style* strategy was used. This format is modelled as the popular game show Jeopardy, including various challenges with different scores. Users must accumulate as many points as possible in a scheduled time, addressing the challenges and finding the right answer. The game's objective is to reduce the learning process time, i.e., facilitate the knowledge acquisition from end users using practical tasks. At the same time, this approach allows developers to speed up the validation cycles with respect to traditional innovation processes.

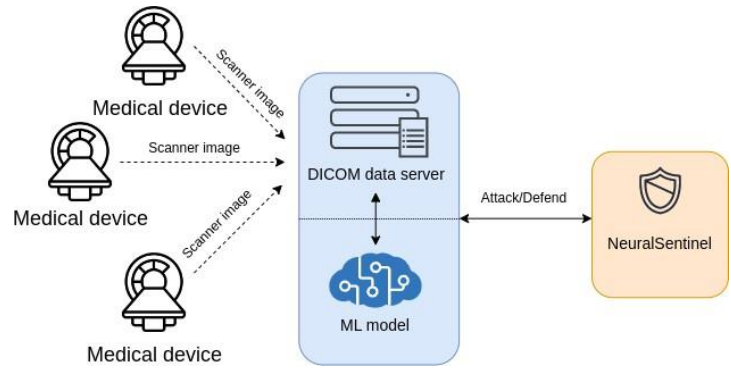

Fig. 6: Healthcare Use Case schema

The first KINAITICS hackathon was composed of 17 challenges aimed at testing 4 different tools. More specifically, 6 of them were directly related to NS. The challenges were self-explanatory, that is, they were designed in a way that any person without expert knowledge about the tool and its application domain (Healthcare) could understand the activity, the context and the required goal. For this reason, as shown in Figure 7, each challenge was detailed with the referred Use Case (*UC*), how to access the tool (*Tool*), a general description of the topics covered, i.e., what the user will learn from addressing the challenge (*Description*), the effective objective of the challenge, i.e., the task to perform (*Goal*) and finally, indications on how to enter and what format to use in the answers box (*Result*).

The 6 NS challenges aimed to exploit and combine all the attack and defence mechanisms implemented in the tool. Figure 7 shows an example related to fine-tuning the parameters of the Basic Iterative Method (BIM) attack algorithm. In this case, the user needed to run several BIM attacks, changing the epsilon and steps values, in order to find the best couple of values able to maximize the grade metric.

### 4.3  Discussion

The hackathon event saw 20 registered participants from 6 European Countries, as shown in Table 1. Participants were both internal (project members) and external end-users and stakeholders, with different profiles (junior, senior, but also practitioners and students) and different domain knowledge (Cybersecurity, Consulting, System Admin, Health Researcher, Software Engineers, etc.). The event lasted four hours, two of which were dedicated to completing all the challenges of all tools.

Fig. 7: Hackathon challenge example

As a general output, participants noted that the given time for completing the activities felt limited. On the other hand, in relation to NS validation, participants suggested that providing brief training on basic concepts related to the functioning of attacks and defence methods in machine learning models could be beneficial. This would help level the initial knowledge of participants and facilitate greater engagement and understanding of the proposed tasks. Furthermore, they found the tool quite user-friendly, i.e., intuitive and self-explanatory. However, a potential area for improvement lies in the visualization, refining specific views and introducing additional graphs to enhance the model explainability and user experience.

It is also important to point out that the involvement of 20 individuals cannot be used as a full and complete validation approach. However, it was useful to detect the benefits and limitations of the tool and to have an initial validation of the activities. In that sense, for future iterations, an effort will be made to involve a larger number of participants, providing an even more comprehensive and diversified perspective on the effectiveness of the hackathon in different contexts.

In summary, the initial validation of the hackathon resulted in high participant satisfaction, while also identifying areas for improvement, such as time management, the inclusion of introductory training and tool usability.

| Hackathon participants statistics | | | | | | |
|---|---|---|---|---|---|---|
| Profile | Number | Country | Number | Country | KINAITICS member? | Number |
| **Project Manager** | 1 | **Austria** | 1 | **Researcher** | YES | 6 |
| **CEO** | 1 | **France** | 2 | **Center** | NO | 1 |
| **Researcher** | 9 | **Greece** | 3 | **University** | YES | 0 |
| **PhD student** | 2 | **Italy** | 9 | | NO | 3 |
| **Post-Doc Researcher** | 1 | **Portugal** | 1 | **End user** | YES | 6 |
| **Software Engineer** | 2 | **Portugal** | 1 | | NO | 4 |
| **Consulting** | 1 | **Spain** | 4 | | | |
| **System Admin** | 1 | | | | | |
| **Engineer** | 1 | | | | | |
| **Cyber-Sec Specialist** | 1 | | | | | |
| **Total** | 20 | - | 20 | - | | 20 |

Table 1: Statistic of Hackathon participants

## 5 Conclusion

This work presents the NeuralSentinel tool that allows users to familiarize themselves with the AI security field. Concretely, it gives the opportunity to analyze their artificial neural networks and how robust and trustworthy their models are. Cybersecurity in AI systems is a novel field in research, even though those models are gaining more and more importance in human lives. However, people are becoming aware of that casuistry and are getting interested in how trustworthy their model is. This idea motivates the generation of this tool that will help those first users understand how their models work and check the robustness of their models. The tool was successfully validated in a Hackathon event organized by the KINAITICS project. The tool was used in a Healthcare scenario with the aim to attack and defend an ML system able to diagnose cancer or no cancer from skin breast images.

In the future, the NeuralSentinel tool will be extended, incorporating more attack/defence mechanisms, such as those mentioned in Section 2. At the same time, will be interesting to improve the XAI module, including analysis performed through frameworks like SHAP or LIME. The extension will allow NeuralSentinel to generate an improved mode assessment of the users' models, carry out a more exhaustive analysis, give more information to the users about their model, and finally, generate more trustworthiness in the users about the robustness of the model. Furthermore, the idea is to incorporate feedback gathered from end users to improve the usability and the layout of the tool.

# References

1. K. E. Henry, R. Kornfield, A. Sridharan, R. C. Linton, C. Groh, T. Wang, A. Wu, B. Mutlu, and S. Saria, "Human–machine teaming is key to ai adoption: clinicians' experiences with a deployed machine learning system," *NPJ digital medicine*, vol. 5, no. 1, p. 97, 2022.
2. S. Huang, N. Papernot, I. Goodfellow, Y. Duan, and P. Abbeel, "Adversarial attacks on neural network policies," 2017.
3. P.-J. Kindermans, K. Schütt, K.-R. Müller, and S. Dähne, "Investigating the influence of noise and distractors on the interpretation of neural networks," 2016.
4. S. G. Finlayson, H. W. Chung, I. S. Kohane, and A. L. Beam, "Adversarial attacks against medical deep learning systems," 2019.
5. N. Akhtar and A. Mian, "Threat of adversarial attacks on deep learning in computer vision: A survey," *IEEE Transactions on Image Processing*, vol. 6, pp. 14 410–14 430, 2018.
6. C. Szegedy, W. Zaremba, I. Sutskever, J. Bruna, D. Erhan, I. Goodfellow, and R. Fergus, "Intriguing properties of neural networks," in *International Conference on Learning Representations*, 2014.
7. I. J. Goodfellow, J. Shlens, and C. Szegedy, "Explaining and harnessing adversarial examples," *arXiv preprint arXiv:1412.6572*, 2014.
8. A. Kurakin, I. J. Goodfellow, and S. Bengio, "Adversarial examples in the physical world," in *5th International Conference on Learning Representations, ICLR 2017 - Workshop Track Proceedings*, 2019, pp. 1–14.
9. A. Madry, A. Makelov, L. Schmidt, D. Tsipras, and A. Vladu, "Towards deep learning models resistant to adversarial attacks," in *International Conference on Learning Representations*, 2018.
10. N. Papernot, P. McDaniel, S. Jha, M. Fredrikson, Z. B. Celik, and A. Swami, "The limitations of deep learning in adversarial settings," in *2016 IEEE European symposium on security and privacy (EuroS&P)*. IEEE, 2016, pp. 372–387.
11. R. Wiyatno and A. Xu, "Maximal jacobian-based saliency map attack," 2018.
12. S. Moosavi-Dezfooli, A. Fawzi, and P. Frossard, "Deepfool: A simple and accurate method to fool deep neural networks," in *2016 IEEE Conference on Computer Vision and Pattern Recognition (CVPR)*, 2016, pp. 2574–2582.
13. X. Echeberria-Barrio, A. Gil-Lerchundi, I. Goicoechea-Telleria, and R. Orduna-Urrutia, "Deep learning defenses against adversarial examples for dynamic risk assessment," in *13th International Conference on Computational Intelligence in Security for Information Systems (CISIS 2020)*, 2021.
14. N. Papernot, P. McDaniel, X. Wu, S. Jha, and A. Swami, "Distillation as a defense to adversarial perturbations against deep neural networks," in *2016 IEEE symposium on security and privacy (SP)*. IEEE, 2016, pp. 582–597.
15. H. Lee, H. Bae, and S. Yoon, "Gradient masking of label smoothing in adversarial robustness," *IEEE Access*, vol. 9, pp. 6453–6464, 2021.
16. S. M. Lundberg and S.-I. Lee, "A unified approach to interpreting model predictions," *Advances in neural information processing systems*, vol. 30, 2017.
17. M. T. Ribeiro, S. Singh, and C. Guestrin, "" why should i trust you?" explaining the predictions of any classifier," in *Proceedings of the 22nd ACM SIGKDD international conference on knowledge discovery and data mining*, 2016, pp. 1135–1144.
18. A. Shrikumar, P. Greenside, and A. Kundaje, "Learning important features through propagating activation differences," in *International conference on machine learning*. PMLR, 2017, pp. 3145–3153.
19. M. D. Zeiler and R. Fergus, "Visualizing and understanding convolutional networks," in *Computer Vision–ECCV 2014: 13th European Conference, Zurich, Switzerland, September 6-12, 2014, Proceedings, Part I 13*. Springer, 2014, pp. 818–833.

20. B. Zhou, A. Khosla, A. Lapedriza, A. Oliva, and A. Torralba, "Learning deep features for discriminative localization," in *Proceedings of the IEEE conference on computer vision and pattern recognition*, 2016, pp. 2921–2929.
21. J. H. Friedman, "Greedy function approximation: a gradient boosting machine," *Annals of statistics*, pp. 1189–1232, 2001.
22. J. B. Alex Goldstein, Adam Kapelner and E. Pitkin, "Peeking inside the black box: Visualizing statistical learning with plots of individual conditional expectation," *Journal of Computational and Graphical Statistics*, vol. 24, no. 1, pp. 44–65, 2015. [Online]. Available: https://doi.org/10.1080/10618600.2014.907095
23. X. Echeberria-Barrio, A. Gil-Lerchundi, J. Egana-Zubia, and R. Orduna-Urrutia, "Understanding deep learning defenses against adversarial examples through visualizations for dynamic risk assessment," *Neural Computing and Applications*, vol. 34, no. 23, pp. 20 477–20 490, 2022.
24. Z. Wang, A. C. Bovik, H. R. Sheikh, and E. P. Simoncelli, "Image quality assessment: from error visibility to structural similarity," *IEEE transactions on image processing*, vol. 13, no. 4, pp. 600–612, 2004.
25. K. Simonyan and A. Zisserman, "Very deep convolutional networks for large-scale image recognition," *arXiv preprint arXiv:1409.1556*, 2014.
26. V. Švábenskỳ, P. Čeleda, J. Vykopal, and S. Brišáková, "Cybersecurity knowledge and skills taught in capture the flag challenges," *Computers & Security*, vol. 102, p. 102154, 2021.